%% file: main.tex
\documentclass{article}
\usepackage{iclr2027_conference}
\iclrfinalcopy

\usepackage{fontspec}
\usepackage{amsmath,amssymb,amsfonts}
\usepackage{booktabs}
\usepackage{array}
\usepackage{tabularx}
\usepackage{enumitem}
\usepackage{xspace}
\usepackage{xcolor}
\usepackage{graphicx}
\usepackage{wrapfig}
\usepackage{float}
\usepackage{placeins}
\usepackage{url}
\usepackage{natbib}
\usepackage{hyperref}
\hypersetup{hidelinks}

\definecolor{aopsdblue}{HTML}{28658F}
\definecolor{aopsdamber}{HTML}{D18B1F}
\definecolor{aopsdred}{HTML}{B84A42}
\definecolor{aopsdtrack}{HTML}{E8EDF1}

\newcommand{\ours}{\textsc{AOPSD}\xspace}
\newcommand{\sg}{\operatorname{sg}}

\newcommand{\epochvalues}[3]{%
  \makebox[1.85em][r]{#1}\kern.12em/\kern.12em%
  \makebox[1.85em][r]{#2}\kern.12em/\kern.12em%
  \makebox[1.85em][r]{#3}}

\title{Learning from a Thoughtful Teacher:\\
Adaptive On-Policy Self-Distillation\\
for Mathematical Reasoning}
\author{%
\textbf{Jiacheng Du}$^{1,*}$ \quad \textbf{Weiwei Xie}$^{2,*}$ \quad \textbf{Tianyi Du}$^{3}$ \\
\textbf{Shaoxiong Guo}$^{3}$ \quad \textbf{Qibing Ren}$^{2,\dagger}$ \quad \textbf{Jiaheng Zhang}$^{1,\dagger}$ \\[0.4em]
{\normalfont\small $^1$National University of Singapore} \\
{\normalfont\small $^2$Shanghai Jiao Tong University} \\
{\normalfont\small $^3$Shanghai Artificial Intelligence Laboratory}
}

\begin{document}

\makeatletter
\@namedef{r@app:motivation}{{A}{}{}{}{}}
\@namedef{r@app:dag-construction}{{B}{}{}{}{}}
\@namedef{r@app:implementation}{{C}{}{}{}{}}
\@namedef{r@app:training-cost}{{D}{}{}{}{}}
\@namedef{r@app:training-prompts}{{E}{}{}{}{}}
\@namedef{r@app:full-results}{{F}{}{}{}{}}
\@namedef{r@app:nll-semantic-audit}{{F.3}{}{}{}{}}
\@namedef{r@app:qwen35-2b-results}{{H}{}{}{}{}}
\@namedef{r@tab:component_ablation}{{6}{}{}{}{}}
\@namedef{r@tab:continuation_ablation}{{7}{}{}{}{}}
\makeatother
\raggedbottom
\maketitle
\lhead{}
\renewcommand{\headrulewidth}{0pt}
\begingroup
\renewcommand{\thefootnote}{}
\footnotetext{%
$^*$Equal contribution. \quad $^\dagger$Corresponding authors.\\
Emails: Jiacheng Du (\texttt{jiachengdu@u.nus.edu}); Weiwei Xie (\texttt{xiewwee11@sjtu.edu.cn}).}
\endgroup

\begin{abstract}
On-policy self-distillation (OPSD) trains a question-only student with token-level feedback from a
teacher given training-only privileged information (PI). OPSD therefore provides dense, on-policy supervision, and is free of a larger external teacher, but its effectiveness rests on how PI
is designed and utilized. Our preliminary diagnostics suggest a significant gap between teacher
utility and student learnability, where a small fraction of high-disagreement tokens dominate the distillation signal, and short teacher continuations at these positions further expose more explicit PI leakage than transferable correction cues, indicating a strong intent on injecting PI-conditioned shortcuts. We propose Adaptive
On-Policy Self-Distillation (\ours{}), which adapts what information the teacher receives and how
strongly its feedback influences learning. \ours{} encodes each solution as a reasoning
DAG, orders problems by the student’s evolving capability, and reveals only
the affordable subgraph and its next frontier as PI. For high-disagreement tokens, \ours{} utilizes short teacher continuations as probes to encourage useful guidance while mitigating PI-conditioned shortcuts among teacher supervisions. On HMMT25, AIME24, AIME25, and BRUMo25, \ours{} achieves 72.5\% Pass@8, which is 6.7 percentage points above OPSD and 4.2 above the strongest competing baseline while reducing 15 percentage points of training time at lower cost.
\end{abstract}

\input{sections/intro}
\input{sections/related}
\input{sections/motivation}

\input{sections/method}

\input{sections/experiments}
\input{sections/conclusion}
\bibliographystyle{iclr2027_conference}
\bibliography{refs}

\end{document}

%% file: sections/intro.tex
\section{Introduction}
\label{sec:intro}

Mathematical problem solving is a demanding test of language-model reasoning: a solver must preserve
constraints, choose among plausible routes, and revise intermediate claims over long trajectories. Teaching mathematical reasoning requires more than supplying a solution: the student must learn
to make the underlying decisions on its own. Supervised fine-tuning (SFT) supplies dense targets, but these targets come
from fixed solution traces rather than the evolving student's attempts
\citep{yue2024mammoth,toshniwal2025openmath}. Reinforcement learning with verifiable rewards
(RLVR) trains on those attempts, but a final-answer reward does not identify which intermediate
choices should change \citep{shao2024deepseekmath,guo2025deepseekr1,yu2025dapo}. The challenge is to provide dense supervision at the
reasoning states the student actually visits.

On-policy distillation addresses this challenge by asking a teacher to score the student's own
trajectories \citep{agarwal2024gkd,gu2024minillm}. It provides token-level feedback at sampled
prefixes, but requires a suitable teacher with accessible probabilities and a compatible output
space \citep{zhang2024dualspace,boizard2025cross}. On-policy self-distillation (OPSD) obtains this
teacher from the same base model by giving it a richer context \citep{zhao2026opsd}: the student
sees only the question, while the teacher also receives training-only privileged information
(PI), such as a verified solution. PI thus supplies the teacher's informational advantage without
requiring a larger external teacher.

This contextual advantage introduces a different teaching problem: information that helps a
teacher solve a problem may not help a student learn to solve it unaided. We distinguish
\emph{conditional utility}, the benefit of PI while it remains in context, from
\emph{question-only transfer}, the distilled student's performance when PI is absent. A teacher
can draw directly on a supplied derivation or conclusion, whereas the student must learn to
produce the reasoning that supports it. 

Our preliminary diagnostics motivate treating utility and transfer as separate criteria
(Section~\ref{sec:motivation}). Full solutions, plans, and subproblem decompositions significantly improve the teacher's utility, yet the corresponding question-only students score below the reported no-PI
reference. At the token level, a small high-disagreement tail accounts for a large share of the
absolute teacher--student log-probability gap and
 dominates the distillation signal. Short teacher continuations at these positions
expose both grounded
corrections and explicit PI leakage, which motivates looking beyond disagreement magnitude when deciding
how to use privileged feedback.

We propose Adaptive On-Policy Self-Distillation (\ours{}), which adapts two teaching decisions to
the evolving student: \emph{what information the teacher receives} and \emph{how strongly its
feedback influences learning}. Specifically, we first identify \emph{reasoning
DAGs as privileged information}, where solutions are compressed into prerequisite-linked mathematical states and can be decomposed without breaking their logic. Then, we design an \emph{adaptive PI disclosure} mechanism. According to the current student's capability, \ours{} adaptively discloses the affordable subgraph as PI for every trained problem, and dynamically reorders the curriculum during the training process. Third, we apply \emph{continuation-based selective distillation} for high-disagreement tokens, where the teacher generates short continuations and the student scores them via negative log-likelihood loss, in order to recognize and follow grounded guidance while mitigating the PI-conditioned bias from the teacher.

Our contributions are:
\begin{itemize}[leftmargin=1.4em,itemsep=1pt,topsep=2pt]
\item We expose a utility–transfer gap in mathematical OPSD: PI can make a substantially stronger
conditional teacher while the resulting student underperforms the reported no-PI condition. We
localize this failure to a small high-disagreement tail and use short continuations to diagnose
whether its dominant preferences correct the problem or expose PI-conditioned behavior.

\item We propose Adaptive
On-Policy Self-Distillation (\ours{}), which adapts what information the teacher receives and how
strongly its feedback influences learning. \ours{} encodes each solution as a reasoning
DAG, updates the curriculum based on the student’s evolving capability, and reveals only
the affordable subgraph and its next frontier as PI. For high-disagreement tokens, \ours{} utilizes short teacher continuations as probes to encourage useful guidance while mitigating PI-conditioned shortcuts among teacher supervision.
\item On four competition-mathematics benchmarks, \ours{} achieves the highest final Pass@8 on every benchmark and a 72.5\% macro-average, exceeding OPSD by 6.7~pp and the strongest competing baseline by 4.2~pp. Compared to the OPSD baseline, our method has lower theoretical memory usage and reducing 15.6 pp training time in empirical tests.
\end{itemize}

%% file: sections/related.tex
\section{Related Work}
\label{sec:related}

\paragraph{Post-training techniques for mathematical reasoning.}
Mathematical post-training draws supervision from worked solutions and verifiable outcomes.
MAmmoTH combines chain-of-thought and program-of-thought rationales \citep{yue2024mammoth},
MetaMath expands mathematical questions through bootstrapping \citep{yu2024metamath}, and
OpenMathInstruct-2 scales synthetic solution data \citep{toshniwal2025openmath}.
These SFT methods provide dense targets on fixed traces. DeepSeekMath, DeepSeek-R1, and DAPO
instead optimize sampled attempts using verifiable rewards
\citep{shao2024deepseekmath,guo2025deepseekr1,yu2025dapo}, but outcome supervision does not directly
identify which intermediate decisions should change. \ours{} uses reference solutions as context
for dense feedback on the student's own attempts, with disclosure tied to its demonstrated progress.

\paragraph{On-policy distillation.}
OPD moves teacher supervision onto prefixes the student actually visits.
GKD learns from self-generated sequences with flexible divergence objectives \citep{agarwal2024gkd},
while MiniLLM uses reverse KL to avoid overestimating low-probability teacher outputs
\citep{gu2024minillm}. Entropy-Aware On-Policy Distillation (EOPD) augments reverse-KL OPD with
forward KL at positions of high teacher entropy, preserving probability mass across plausible
alternatives, and adapting distribution matching to teacher uncertainty\citep{jin2026eopd}. However, OPD requires direct matching of token distributions from open-source teachers, and different output vocabularies introduce an additional alignment problem
\citep{zhang2024dualspace,boizard2025cross}. Therefore, our work mainly focuses on self-distillation and studies how to use the contextual advantage of a teacher with the same initialization.

\paragraph{On-policy self-distillation.} OPSD conditions the teacher on a reference solution and matches its predictions along
student-generated trajectories \citep{zhao2026opsd}. OP$^2$SD replaces the paired reference with
another problem's worked example, showing that context-induced teacher behavior can preserve
substantial gains without target-specific solutions \citep{ichihara2026op2sd}. DASH adapts
token-level supervision by aggregating local divergences over horizons determined by their
variation along the rollout \citep{hou2026dash}. These methods establish the value of contextual
teaching and adaptive supervision, but do not jointly align the disclosed information and the
feedback weights with the student's current reasoning capability. \ours{} couples these decisions:
DAG-based progress controls disclosure up to the next dependency frontier and orders the
problem curriculum, while teacher continuations scored by the question-only student determine
feedback strength.

%% file: sections/motivation.tex
\section{Motivation: When Privileged Utility Fails to Transfer}
\label{sec:motivation}

A useful teacher must do more than solve a problem with extra information: its feedback must
help a student that will not receive that information. We distinguish \emph{conditional utility},
the benefit of PI to a solver while it remains in context, from \emph{question-only transfer}, the
performance of the student after distillation with PI removed. The following preliminary
diagnostics examine this distinction at two levels: final outcomes and local teacher--student
disagreements.

\subsection{Observation 1: A good privileged solver can be a bad teacher}
\label{sec:obs-solver-teacher}

Our initial comparison uses Qwen3-4B-Instruct-2507 and 725 competition-mathematics problems with verified
solutions. We derive three teacher contexts from each solution: the full solution, a solution
plan, and a subproblem decomposition. Conditional utility is measured by solving these problems
with each context; transfer is assessed by evaluating the distilled question-only
students on AIME~2025. Appendix~\ref{app:motivation} provides the diagnostic settings and examples of the PI formats.

\begin{figure}[t]
  \centering
  \includegraphics[width=0.9\linewidth]{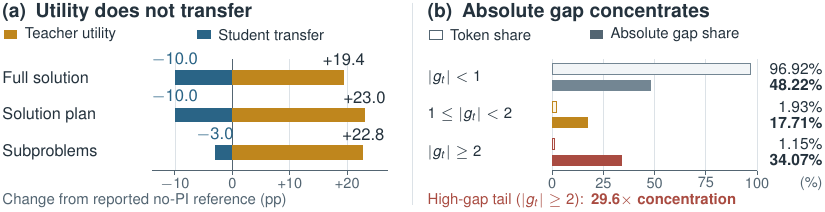}
  \caption{\textbf{Preliminary diagnostics of PI utility and transfer.}
  (a) Change relative to the baseline model: every PI
makes the privileged teacher substantially stronger (amber) while the resulting question-only student
gets weaker (blue). (b) Token share and share of total absolute sampled-token log gap, grouped by
  $|g_t|$.}
  \label{fig:motivation-diagnostics}
\vspace{-4mm}
\end{figure}

All three PI formats substantially improve conditional solving in this diagnostic. Pass@8 rises
from $554/725$ ($76.4\%$) with the question alone to $695/725$ ($95.9\%$) with the full solution,
$721/725$ ($99.4\%$) with a plan, and $719/725$ ($99.2\%$) with subproblems. Yet the corresponding
question-only students score $56\%$, $56\%$, and $63\%$ after vanilla OPSD training, below the reported $66\%$ baseline (Figure~\ref{fig:motivation-diagnostics}a). We further examined students' rollouts on AIME25. A striking phenomenon was that students, seeing only the questions during the test, generated statements like \textquotedblleft{}the given solution", which would only be reasonable given PI as prompts, which suggests that some PI-conditioned behaviors may have been distilled and transferred to the question-only student.

To inspect the feedback behind these outcomes, we measure the teacher-student disagreement along question-only
student rollouts. For a problem $x$, rollout $y$, and privileged context $P$, let $p$ denote the
student and $q^P$ the PI-conditioned teacher. The sampled-token log-probability gap is
\begin{equation}
g_t(P)=\log q^{P}(y_t\mid x,y_{<t})-\log p(y_t\mid x,y_{<t}).
\label{eq:motivation-gap}
\end{equation}
In the collected rollouts, only $3.08\%$ of tokens satisfy $|g_t|\geq1$, but they account for
$51.78\%$ of $\sum_t|g_t|$. The $|g_t|\geq2$ subset contains $1.15\%$ of tokens and $34.07\%$
of the absolute gap (Figure~\ref{fig:motivation-diagnostics}b). This concentration motivates
closer inspection of high-gap positions. 

\subsection{Observation 2: Teacher introduces shortcuts instead of useful guidance}
\label{sec:obs-heavy-tail}

A large token-level gap provides little context about the behavior that follows it. The sampled
token may be an ordinary connector such as \texttt{But}. In a separate diagnostic using a
reasoning DAG as PI, we select non-overlapping events with $|g_t|\geq2$, hold the student's token
$y_t$ fixed, and let the privileged teacher continue greedily. Each continuation contains $y_t$ and up
to $k-1$ new tokens, exposing a short branch after the original disagreement.

\begin{table}[t]
  \centering
  \caption{\textbf{A short continuation makes high-disagreement intent legible.}
    Cumulative counts of recognizable correction cues and explicit PI leakage within a total continuation of $k$ tokens.}
  \label{tab:continuation-intent}
  \begingroup
  \definecolor{AOPSDInk}{HTML}{24323D}
  \definecolor{AOPSDCorrection}{HTML}{2A6486}
  \definecolor{AOPSDLeakage}{HTML}{A94B41}
  \color{AOPSDInk}
  \fontsize{9}{10.5}\selectfont
  \setlength{\tabcolsep}{5pt}
  \renewcommand{\arraystretch}{1.12}
  \setlength{\aboverulesep}{2.2pt}
  \setlength{\belowrulesep}{2.2pt}
  \begin{tabularx}{\linewidth}{@{}c
    >{\color{AOPSDCorrection}\bfseries}r
    >{\hsize=0.92\hsize\linewidth=\hsize\raggedright\arraybackslash\itshape}X
    >{\color{AOPSDLeakage}\bfseries}r
    >{\hsize=1.08\hsize\linewidth=\hsize\raggedright\arraybackslash\itshape}X@{}}
    \toprule[0.7pt]
    $k$ & \multicolumn{2}{c}{\color{AOPSDCorrection}\bfseries Correction cues}
      & \multicolumn{2}{c}{\color{AOPSDLeakage}\bfseries Explicit PI references} \\
    \cmidrule(lr){2-3}\cmidrule(l){4-5}
    (tokens) & \multicolumn{1}{r}{Count} & \multicolumn{1}{l}{Representative excerpt}
      & \multicolumn{1}{r}{Count} & \multicolumn{1}{l}{Representative excerpt} \\
    \midrule[0.4pt]
    2  & 3  & an error
       & 8   & original DAG \\
    4  & 7  & Not quite---
       & 70  & checking the private reasoning DAG \\
    6  & 22 & the density argument must be flawed
       & 153 & go back to the private reasoning DAG \\
    8  & 31 & Or maybe we made a mistake?
       & 238 & This matches the reasoning in the DAG \\
    16 & 64 & Wait---not necessarily.
       & 507 & earlier in the DAG, it says \ldots{} \\
    \bottomrule[0.7pt]
  \end{tabularx}
  \endgroup
  \vspace{-4mm}
\end{table}

Longer windows reveal both correction cues and explicit references to the private DAG
(Table~\ref{tab:continuation-intent}). A phrase such as ``an error'' suggests an attempted
correction; a reference to ``the private reasoning DAG'' explicitly invokes context unavailable
to the student, which implies that magnitude alone does not characterize trusted signal from the teacher, and teachers' on-policy guidance is actually more about introducing shortcuts than correcting errors.

%% file: sections/method.tex
\section{\ours{}: Teaching at the Learner's Frontier}
\label{sec:method}

\subsection{Overview}
\label{sec:overview}

Vanilla OPSD fixes the teacher's privileged context and applies the same distillation rule across
sampled tokens. The diagnostics in Section~\ref{sec:motivation} suggest that neither choice should
be taken for granted: useful PI may fail to transfer, and a large teacher–student gap alone does not reveal which direction the teacher is encouraging. \ours{} makes both choices relative to the current student
rollout. It controls what part of PI the teacher receives and, at high-gap positions, how strongly
the teacher’s feedback affects the student.

\begin{figure}[t]
  \centering
  \includegraphics[width=0.9\linewidth]{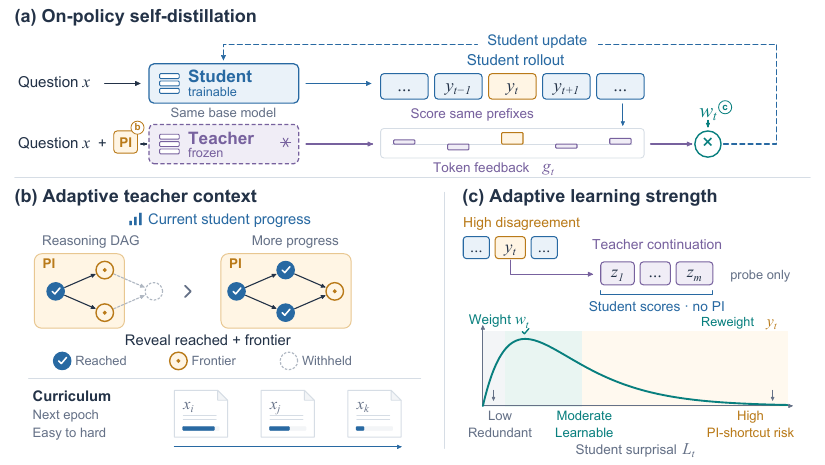}
  \caption{\textbf{\ours{}: adapt what to reveal and how strongly to learn.}
  (a) Teacher and student share an initialization, and the teacher scores the student's question-only trajectory using privileged information.
(b) Based on the student’s current capability, the DAG is adaptively disclosed, and all questions are presented through curriculum learning in order of difficulty.
(c) High-gap tokens trigger teacher continuations scored by the question-only student. Band-pass weighting suppresses redundant and potentially PI-dependent guidance while
retaining the learnable and useful guidance.}
  \label{fig:aopsd-overview}
\vspace{-2mm}
\end{figure}

The method links three components (Figure~\ref{fig:aopsd-overview}). A reasoning DAG represents
the reference solution as mathematical checkpoints with explicit prerequisites
(Section~\ref{sec:dag}). The checkpoints established by the student's rollout determine both
the subgraph disclosed to the teacher and the problem's position in a curriculum
(Section~\ref{sec:disclosure}). At high-disagreement tokens, a short teacher continuation is
scored by the question-only student to set a distillation weight
(Section~\ref{sec:selective}). Together, these components adapt the teacher’s privileged view and feedback strength to the student’s current state.

\subsection{Reasoning DAGs as privileged information}
\label{sec:dag}

Selective disclosure requires a representation that can be partially revealed without breaking the dependencies of a solution. An
arbitrary text prefix can end midway through an inference, whereas a dependency graph makes
its prerequisites explicit. We represent one verified solution route by a nonempty reasoning
DAG $\mathcal G_i=(\mathcal V_i,\mathcal E_i)$. Each node is a mathematical checkpoint, corresponding to an intermediate state required by the solution; an edge
$u\to v$ states that $u$ is an immediate prerequisite of $v$. Fan-out represents parallel
arguments, and fan-in records where those arguments must be combined.

\begin{figure}[t]
  \centering
  \includegraphics[width=0.9\linewidth]{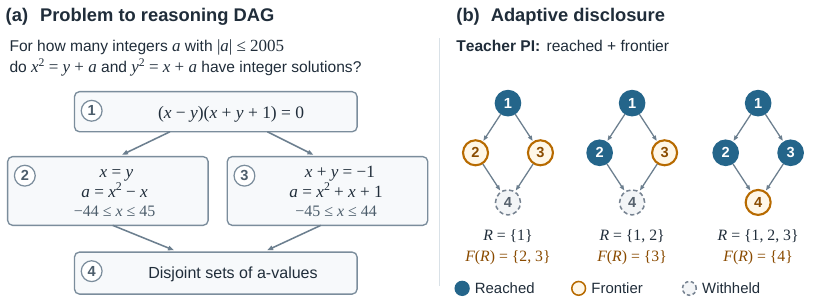}
  \caption{\textbf{A reasoning DAG reveals one complete dependency frontier at a time.}
  (a) A verified route is compressed into route-defining mathematical checkpoints; edges encode
  immediate prerequisites. 
  (b) For a question-only rollout, reached states $R$ are blue, immediately enabled states $F(R)$
  are orange, and later states are gray. As more prerequisites are established, disclosure advances
  by one dependency layer rather than by an arbitrary text prefix.}
  \label{fig:dag-example}
\vspace{-4mm}
\end{figure}

We construct the graphs once, offline. A producer converts a verified solution into a dense
dependency graph, which an independent judge reviews for fidelity and completeness. A second producer compresses the accepted graph into key checkpoints and
is instructed to remove the terminal-answer layer. Appendix~\ref{app:dag-construction} details the construction procedure and gives examples of the resulting DAGs.

The graph serves two roles: it supplies decomposable units of PI and a reference against which to assess a
student's completeness and capability (Figure~\ref{fig:dag-example}). Its checkpoints make both disclosure and progress assessment explicit at the level of one
verified route.

\subsection{Adaptive PI disclosure}
\label{sec:disclosure}

\paragraph{Measure progress along the reference route.}
OPSD already generates a question-only response before teacher feedback. We reuse that rollout for
an inference-only alignment against $\mathcal G_i$. Let $R_i(y)$ be the largest set of checkpoints
judged to be established by $y$ after removing any checkpoint whose prerequisites are absent. By
construction, $R_i(y)$ is prerequisite closed. We define
\begin{equation}
r_i(y)=\frac{|R_i(y)|}{|\mathcal V_i|}\in[0,1].
\label{eq:dag-progress}
\end{equation}
$r_i(y)$ measures progress along the reference route, giving a finer estimate of the current problem’s
difficulty for the trained student model than final correctness, which is what both adaptations need. In practice, we prompt student model as the judger for correctness estimation.

\paragraph{Local adaptation: disclose the reached subgraph and its frontier.}
The teacher sees what the student has
reached, plus the next checkpoints whose prerequisites are already reached:
\begin{equation}
F_i(R)=\{v\in\mathcal V_i\setminus R:\operatorname{pa}(v)\subseteq R\},\qquad
P_i(y)=\phi\!\left(\mathcal G_i[R_i(y)\cup F_i(R_i(y))]\right),
\label{eq:local-disclosure}
\end{equation}
where $\operatorname{pa}(v)$ denotes the parents of $v$, $\mathcal G_i[S]$ is the subgraph induced
by $S$, and $\phi$ serializes that subgraph as teacher context.  A weak
student unlocks little beyond the roots, and a stronger student unlocks a deeper subgraph
(Figure~\ref{fig:dag-example}b). The boundary is deliberate on both sides. Revealing more would
place the teacher's preferences on states the student has shown no route to, which is the failure
of Section~\ref{sec:motivation}; revealing less would withhold the one transition the student is
ready to make. This keeps PI near the next missing step, rather than adding
downstream details whose prerequisites the student has not yet established.

\paragraph{Global adaptation: update problem curriculum with problem competence.}
Averaged over a problem's attempts, the same completion is a per-problem competence. We initialize an exponential moving average from $K$ question-only attempts and
update it at epoch boundaries:
\begin{equation}
c_i^{(0)}=\frac{1}{K}\sum_{y\in\mathcal Y_i^{(0)}}r_i(y),\qquad
c_i^{(e+1)}=(1-\lambda)c_i^{(e)}+\lambda\,
\mathbb E_{y\in\mathcal Y_i^{(e)}}[r_i(y)].
\label{eq:dynamic-competence}
\end{equation}
The next epoch visits all problems from higher to lower $c_i$, while bridge batches mix neighboring
coverage strata to soften abrupt transitions. That is, within a single batch, difficult, moderate, and easy questions are mixed in a 1:2:1 ratio relative to the current model's capabilities. Thus, the local and global rules share one measured
quantity: $r_i(y)$ controls how much of a route is disclosed for the current rollout, while $c_i$
controls where the problem appears in the next epoch.

\subsection{Continuation-based selective distillation}
\label{sec:selective}

\begin{figure}[t]
  \centering
  \includegraphics[width=0.8\linewidth]{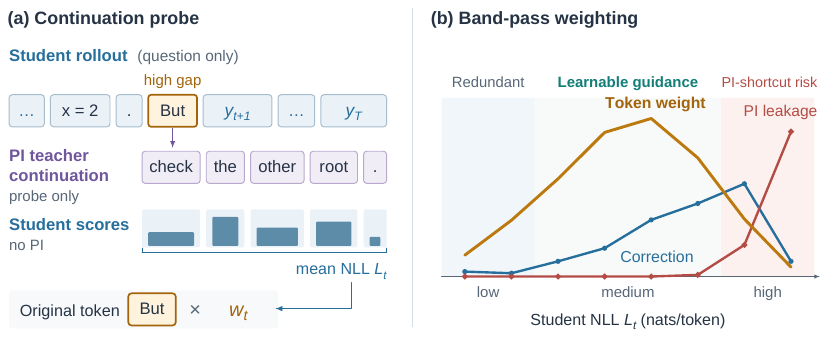}
  \caption{\textbf{Continuation surprisal controls the token's weight.}
  (a) For a large gap, the PI-conditioned teacher generates a continuation after the targeted token. The student then scores it via mean NLL $L_t$.
  (b) The band-pass rule favors intermediate surprisal relative to the two extremes.}
  \label{fig:continuation-weight}
  \vspace{-4mm}
\end{figure}

Disclosure controls the teacher's context, but individual token preferences can still reflect
behavior that is difficult to transfer. We therefore supplement the sampled-token gap with a
short continuation probe. Let $p_{\theta}$ denote the student policy that collected
$y$. At collection time, define
\begin{equation}
g_t^{\mathrm{col}}=
\log q_{\bar\theta}(y_t\mid x_i\oplus P_i(y),y_{<t})
-\log p_{\theta}(y_t\mid x_i,y_{<t}),
\label{eq:opsd-gap}
\end{equation}
where $x_i\oplus P_i(y)$ denotes the teacher prompt containing the problem and disclosed PI.
We probe positions in $\mathcal H(y)=\{t:|g_t^{\mathrm{col}}|\geq\delta\}$ for a threshold
$\delta>0$, where the teacher and student disagree significantly. For each
$t\in\mathcal H(y)$, we choose an anchor $a_t$: retain $a_t=y_t$ for a positive
gap, or let the PI-conditioned teacher choose $a_t$ greedily from $y_{<t}$
for a negative gap. The teacher then generates a greedy suffix
$z^{(t)}=(z_1,\ldots,z_{m_t})$ of at most $k-1$ tokens after the anchor.
The question-only student scores the nonempty suffix, excluding the anchor:
\begin{equation}
L_t=-\frac{1}{m_t}\sum_{j=1}^{m_t}
\log p_{\theta}(z_j\mid x_i,y_{<t},a_t,z_{<j}),
\qquad 1\leq m_t\leq k-1.
\label{eq:continuation-nll}
\end{equation}
Thus $L_t$ measures how surprising the teacher's
short continuation is under the student policy. Our design hypothesis is that highly predictable continuations often add little guidance,
whereas extremely surprising ones may rely on PI unavailable to the student. Intermediate
surprisal offers a candidate balance between novelty and local support. We test this hypothesis with an NLL-blind semantic audit of
15{,}043 training-time continuation events
(Appendix~\ref{app:nll-semantic-audit}).
Useful guidance is enriched at intermediate NLL, whereas 91.35\% of
events with $L_t\geq4$ are labeled as PI leakage. We encode this
hypothesis with a band-pass coefficient:
\begin{equation}
w_t=
\begin{cases}
L_t/\beta_t\times\exp(1 -L_t/\beta_t), & t\in\mathcal H(y)\text{ and }m_t>0,\\
1, & \text{otherwise},
\end{cases}
\qquad \beta_t>0.
\label{eq:continuation-weight}
\end{equation}
We use $\beta_t=\beta_+$ for positive gaps and $\beta_t=\beta_-$ for negative
gaps. The weight vanishes as $L_t\to0$ or $L_t\to\infty$ and peaks at
$L_t=\beta_t$ with value $1$. Thus, $\beta_t$ sets the preferred surprisal
while the maximum weight remains fixed. Untriggered tokens and empty probes
retain unit weight.

\subsection{Training objective}
\label{sec:objective}

\ours{} trains the student with a sampled-token policy-gradient update. The collection-time
gap in Equation~\eqref{eq:opsd-gap} determines the probes. During each actor forward pass, we
form a clipped, weighted advantage using the current student log-probability:
\begin{equation}
\widehat A_t=\operatorname{clip}\!\left(
\sg\!\left[w_t\left(
\log q_{\bar\theta}(y_t\mid x_i\oplus P_i(y),y_{<t})-
\log p_\theta(y_t\mid x_i,y_{<t})\right)\right],-C,C\right),
\label{eq:weighted-advantage}
\end{equation}
where $\sg$ stops gradients and $C>0$ bounds the advantage magnitude. The policy-gradient update
uses $\widehat A_t$ on the original student token $y_t$: a positive advantage favors that token,
and a negative advantage discourages it. The weighted advantage enters a
clipped policy-gradient objective on the sampled responses; the continuation
tokens supply weights rather than additional training targets.
Appendix~\ref{app:implementation} gives the pseudocode, key configuration, and implementation details.

%% file: sections/experiments.tex
\section{Experiments}
\label{sec:experiments}

\subsection{Experimental Setup}
\label{sec:experimental_setup}

\paragraph{Data and evaluation.}
We use the same 725 problems from LIMO-V2~\citep{ye2025limo} as training data for all methods and evaluate on four benchmarks: HMMT25~\citep{flageval2025hmmt} (February), AIME24~\citep{h4_2025aime2024}, AIME25~\citep{testtime2025aime2025}, and BRUMo25~\citep{matharena2025brumo}. We sample eight responses per problem at temperature $0.6$, with at most $12{,}288$ output tokens, and report Pass@8: the percentage of problems with at least one correct response. We compare checkpoints after each of the first three training epochs. ``Avg.'' is the unweighted mean over the four benchmarks for baseline comparisons. All gains are in percentage points (pp). Appendix~\ref{app:full-results} provides complete baseline, component, and sensitivity results and details the evaluation scope.

\paragraph{Baselines and configuration.} We utilize Qwen3-4B-Instruct-2507~\citep{qwen2025qwen3} as the base model for the main evaluations. We additionally evaluate Qwen3.5-2B on the same four benchmarks in Appendix~\ref{app:qwen35-2b-results}.
We compare with the base model before task-specific training, supervised fine-tuning (SFT), GRPO~\citep{shao2024deepseekmath}, and three distillation baselines: OPSD~\citep{zhao2026opsd}, OP$^2$SD~\citep{ichihara2026op2sd}, and DASH~\citep{hou2026dash}. OPSD, OP$^2$SD, and DASH condition the teacher on an example solution. \ours{} uses adaptive DAG disclosure with continuation threshold $\delta=2$, total probe length $k=8$, and sign-specific scales $(\beta_+,\beta_-)=(1,2.5)$. Student rollouts use temperature $1.0$ and teacher probes are greedy. Appendix~\ref{app:implementation} gives the algorithm and key settings; Appendix~\ref{app:training-prompts} gives the prompts.

\subsection{Main Results: Stronger Transfer Throughout Training}
\label{sec:main_results}
\begin{table}[t]
\centering
\caption{\textbf{Overall performance.}
Pass@8 (\%, $\uparrow$). The left block compares the final reported checkpoint; the right block shows the four-benchmark macro-average trajectory. }
\label{tab:main_results}
\small
\setlength{\tabcolsep}{4.4pt}
\renewcommand{\arraystretch}{1.10}
\begin{tabular}{@{}lrrrrrrr@{}}
\toprule
& \multicolumn{4}{c}{Across benchmarks} & \multicolumn{3}{c}{Avg. by epoch} \\
\cmidrule(lr){2-5}\cmidrule(l){6-8}
Method & HMMT25 & AIME24 & AIME25 & BRUMo25 & E1 & E2 & E3 \\
\midrule
Base & 46.7 & 76.7 & 66.7 & 73.3 & 65.9 & 65.9 & 65.9 \\
SFT & 46.7 & 80.0 & 63.3 & 76.7 & 65.0 & 66.7 & 66.7 \\
GRPO & 46.7 & 83.3 & 66.7 & 76.7 & 70.8 & 70.8 & 68.4 \\
OPSD & 43.3 & 76.7 & 66.7 & 76.7 & 63.4 & 68.4 & 65.9 \\
OP$^2$SD & 46.6 & 80.0 & 70.0 & 76.7 & 69.2 & 68.3 & 68.3 \\
DASH & 46.7 & 76.7 & 66.7 & 76.7 & 70.1 & 65.9 & 66.7 \\
\midrule
\textcolor{aopsdblue}{\textbf{AOPSD}} & \textbf{50.0} & \textbf{86.7} & \textbf{73.3} & \textbf{80.0} & \textbf{74.2} & \textbf{72.5} & \textbf{72.5} \\
\bottomrule
\end{tabular}
\vspace{-3mm}
\end{table}

\paragraph{The gains extend across benchmarks.} \ours{} reaches 72.5\% average Pass@8, improving over OPSD by 6.7~pp and over the strongest competing average by 4.2~pp (Table~\ref{tab:main_results}). It also leads every individual benchmark, with gains over the base model of 3.3~pp on HMMT25, 10.0~pp on AIME24, 6.6~pp on AIME25, and 6.7~pp on BRUMo25.
\ours{} achieves 74.2\% in Epoch~1 and retains 72.5\% in both later checkpoints (Table~\ref{tab:main_results}, right), while GRPO finishes at 68.4\% and OPSD at 65.9\%. \ours{} leads or ties the baseline methods in all twelve benchmark--checkpoint comparisons, sustaining its advantage throughout training.

\begin{wraptable}{R}{0.50\textwidth}
  \centering
  \setlength{\abovecaptionskip}{0pt}
  \setlength{\belowcaptionskip}{5pt}
  \caption{\textbf{Empirical test on training time}}
  \label{tab:gradient-time}

  \footnotesize
  \definecolor{costslate}{HTML}{708596}
  \setlength{\tabcolsep}{0pt}
  \renewcommand{\arraystretch}{1.18}

  \newcommand{\costbar}[2]{%
    \makebox[44pt][l]{%
      \rlap{\textcolor{aopsdtrack}{\rule{44pt}{4pt}}}%
      \textcolor{#1}{%
        \rule{\dimexpr44pt*#2/32\relax}{4pt}}}}

  \begin{tabularx}{\linewidth}{@{}
    >{\raggedright\arraybackslash}X
    @{\hspace{5pt}}l
    @{\hspace{3pt}}r
    @{\hspace{7pt}}r@{}}
    \toprule
    Configuration
      & \multicolumn{2}{c}{%
          \shortstack{Time $\downarrow$\\(min/epoch)}}
      & \shortstack[r]{Relative\\time} \\
    \midrule
    Sampled-token PG
      & \costbar{costslate}{24}
      & 24
      & $1.000\times$ \\
    \textcolor{aopsdblue}{\textbf{PG + continuation}}
      & \costbar{aopsdblue}{27}
      & \textcolor{aopsdblue}{\textbf{27}}
      & \textcolor{aopsdblue}{$1.125\times$} \\
    Top-16 forward KL
      & \costbar{costslate}{30}
      & 30
      & $1.250\times$ \\
    Top-32 forward KL
      & \costbar{costslate}{32}
      & 32
      & $1.333\times$ \\
    
    \bottomrule
  \end{tabularx}
  \vspace{-1mm}
\end{wraptable}

\paragraph{Computational cost.}
\ours{} preserves OPSD's dominant computation: student rollout, teacher scoring, and one actor update. Relative to the KL-based OPSD, OP$^2$SD, and DASH, the main concerns on cost are the sampled-token policy-gradient objective (PG) and selective continuation probes. The objective reduces the target-specific loss from $K_d$ terms to one per token and supervision storage from $O(NK_d)$ to $O(N)$; the resulting payload is 89.8\% smaller than Top-16 and 94.9\% smaller than Top-32 KL. Besides, probes require inference only: with a hypothetical 1\% trigger rate and cached-prefix reuse, their analytical overhead is just 2.14\% of the reference distillation cost. In empirical tests, PG keeps selective feedback lightweight: PG with continuation takes 27 minutes per epoch, versus 30 for Top-16 KL and 32 for Top-32 KL (Table~\ref{tab:gradient-time}). This reduces epoch time by 10.0\% and 15.6\%, respectively. Appendix~\ref{app:training-cost} details the timing scope and the computation costs.

\FloatBarrier
\subsection{Ablation Analysis: Context, Curriculum, and Feedback}
\label{sec:component_ablation}
\label{sec:pi_disclosure_ablation}

\begin{wrapfigure}{R}{0.45\textwidth}
  \centering
  \setlength{\parskip}{0pt}
  \includegraphics[
    width=\linewidth,
    trim=7bp 6bp 4bp 4bp,
    clip
  ]{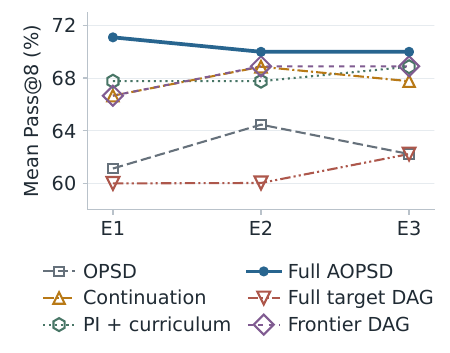}\par
  \setlength{\abovecaptionskip}{4pt}
  \setlength{\belowcaptionskip}{0pt}
  \small
  \caption{\textbf{Ablation study.}
  Continuation and PI + curriculum add one adaptation to OPSD;
  Full AOPSD combines both.
  The DAG variants change context under the OPSD objective.
  Table~\ref{tab:component_ablation} gives all scores.}
  \label{fig:exp-dynamics}
  \vspace{-1cm}
\end{wrapfigure}

Figure~\ref{fig:exp-dynamics} compares six configurations through their average performance across HMMT25, AIME24, and AIME25: component additions to OPSD and DAG-disclosure variants under the OPSD objective. Table~\ref{tab:component_ablation} reports every benchmark--checkpoint score.

\paragraph{Context and feedback adaptation both contribute.}
At Epoch~3, continuation alone raises the average from 62.2\% to 67.8\%, while adaptive PI with curriculum reaches 68.9\%; the full method reaches 70.0\%. Combining the adaptations improves over the stronger individual component by 3.3~pp in Epoch~1 and 1.1~pp in both later epochs. The gains support adapting both the context that induces teacher preferences and the weight those preferences receive in the update.

\paragraph{Selective disclosure matters beyond DAG structure.}
With the OPSD objective fixed, replacing the full target DAG with the student-dependent frontier raises the final average from 62.2\% to 68.9\%. Merely structuring the complete solution therefore does not yield the same transfer as limiting what the teacher sees. Full AOPSD further improves over frontier disclosure by 4.4~pp in Epoch~1 and 1.1~pp at both later checkpoints. 

\begin{figure}[t]
\centering
\includegraphics[width=0.8\linewidth]{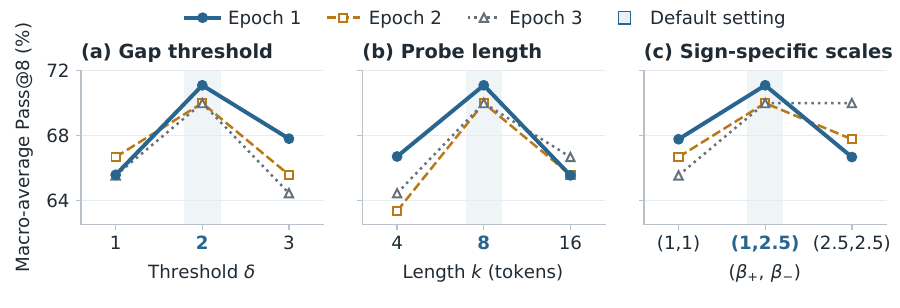}
\caption{\textbf{Hyperparameter analysis.}
Each panel varies one factor around $(\delta,k,\beta_+,\beta_-)=(2,8,1,2.5)$, and Full benchmark scores appear in Table~\ref{tab:continuation_ablation}.}
\label{fig:exp-sensitivity}
\vspace{-3mm}
\end{figure}

\FloatBarrier
\subsection{Hyperparameter Analysis}
\label{sec:continuation_ablation}

\paragraph{Probe the high-gap tail selectively.}
The intermediate threshold $\delta=2$ outperforms $\delta=1$ and $\delta=3$ at every checkpoint, reaching 70.0\% versus 65.5\% and 64.4\% in Epoch~3 (Figure~\ref{fig:exp-sensitivity}a). Broader and narrower triggering both weaken transfer in this sweep: disagreement magnitude identifies where to inspect feedback, but the most extreme cutoff is not best.

\paragraph{A Longer continuation would not help.}
The eight-token budget gives the strongest average at all three checkpoints, exceeding the four- and sixteen-token alternatives by 5.6 and 3.3~pp at Epoch~3 (Figure~\ref{fig:exp-sensitivity}b). This supports using a short local probe: additional teacher generation is useful only insofar as it improves the weight on the original student token. 

\paragraph{Sign-specific scales improve early learning.}
The asymmetric choice $(1,2.5)$ gives the highest average in Epochs~1 and~2; the symmetric $(2.5,2.5)$ choice catches up at Epoch~3, while $(1,1)$ remains at 65.5\% (Figure~\ref{fig:exp-sensitivity}c). The early advantage is consistent with treating reinforcement of the sampled token and redirection toward a teacher alternative differently. The late tie also shows that asymmetry is not necessary to reach the final observed average. 

%% file: sections/conclusion.tex
\section{Conclusion}
\label{sec:conclusion}

On-policy supervision does not by itself make privileged guidance transferable. Our study shows that a teacher can become a stronger conditional solver yet provide feedback that fails to improve question-only reasoning. AOPSD addresses this gap by regulating both the information that induces teacher preferences and the influence those preferences have on learning. Reasoning-DAG progress guides disclosure and curriculum, while student-scored continuation probes calibrate high-disagreement feedback according to its surprisal without privileged information.
Across four benchmarks, AOPSD achieves a final macro-average Pass@8 of 72.5\%, exceeding OPSD by 6.7~pp. Ablations further show that student-dependent disclosure transfers better than exposing the full DAG, and that jointly adapting context and feedback improves the final average over either component alone. Together, these findings suggest a learner-centered principle for privileged self-distillation: use the teacher’s informational advantage to extend the student’s reasoning frontier, rather than substitute for the reasoning the student must learn to perform unaided.